\pdfoutput=1

\documentclass[11pt]{article}

\usepackage{acl}

\usepackage{times}
\usepackage{latexsym}

\usepackage[T1]{fontenc}

\usepackage[utf8]{inputenc}

\usepackage{microtype}

\usepackage{inconsolata}

\usepackage{graphicx}

\usepackage[skip=0pt]{caption}
\usepackage{subcaption}
\usepackage{amsmath}
\usepackage{amssymb}
\usepackage{pifont}

\usepackage{booktabs}
\usepackage{algorithm}
\usepackage{algpseudocode}
\usepackage{algpseudocode}
\usepackage{threeparttable}
\usepackage{multirow}
\usepackage{tabularx}
\usepackage{cleveref}
\usepackage[most]{tcolorbox}
\usepackage{soul}
\usepackage{colortbl}
\usepackage{transparent}
\usepackage{xspace}
\usepackage{paralist}
\usepackage{longtable}
\usepackage[edges]{forest}
\usepackage[textsize=tiny]{todonotes}
\usepackage{listings}
\usepackage{tcolorbox}
\usepackage{verbatim}
\usepackage[normalem]{ulem}
\usepackage{enumitem}
\colorlet{lightCornflowerBlue}{White!80!CornflowerBlue}

\newcommand{\kurier}[1]{{\fontfamily{kurier}\selectfont #1}}

\usepackage{pdflscape}
\usepackage{array}
\usepackage{rotating}
\usepackage{makecell}      
\usepackage{adjustbox}

\crefname{figure}{Fig.}{Figs.}
\crefname{section}{\S}{\S\S}
\crefname{equation}{Eqn.}{Eqns.}
\crefname{appendix}{Appx.}{Appx.}
\crefname{table}{Table}{Tables}

\newcommand{\interalia}{\emph{inter alia}}

\title{\llmformatter: Structuring the Output of Large Language Models}

\author{%
{Darren Yow-Bang Wang}\thanks{~~ Contributed equally.}~, {Zhengyuan Shen}\footnotemark[1], Soumya Smruti Mishra \\
\textbf{Zhichao Xu}, \textbf{Yifei Teng}, \textbf{Haibo Ding} \\
Amazon Web Services \\
\{\texttt{ybwang}, \texttt{donshen}, \texttt{soumish}, \texttt{xzhichao}, \texttt{yifeit}, \texttt{hbding}\}\texttt{@amazon.com}
}

\newcommand{\llmformatter}{\textbf{\kurier{SLOT}}}

\begin{document}
\maketitle
\begin{abstract}

Structured outputs are essential for large language models (LLMs) in critical applications like agents and information extraction. Despite their capabilities, LLMs often generate outputs that deviate from predefined schemas, significantly hampering reliable application development. We present \llmformatter~(\textbf{S}tructured \textbf{L}LM \textbf{O}utput \textbf{T}ransformer), a model-agnostic approach that transforms unstructured LLM outputs into precise structured formats. While existing solutions predominantly rely on constrained decoding techniques or are tightly coupled with specific models, \llmformatter~employs a fine-tuned lightweight language model as a post-processing layer, achieving flexibility across various LLMs and schema specifications. We introduce a systematic pipeline for data curation and synthesis alongside a formal evaluation methodology that quantifies both schema accuracy and content fidelity. Our results demonstrate that fine-tuned Mistral-7B model with constrained decoding achieves near-perfect schema accuracy (99.5\%) and content similarity (94.0\%), outperforming Claude-3.5-Sonnet by substantial margins (+25 and +20 percentage points, respectively). Notably, even compact models like Llama-3.2-1B can match or exceed the structured output capabilities of much larger proprietary models when equipped with \llmformatter, enabling reliable structured generation in resource-constrained environments.
\end{abstract}

\section{Introduction}
\label{sec:intro}

The emergence of large language models (LLMs) has led to a wide range of applications that leverage their natural language understanding capabilities. In such applications, developers are often required to carefully craft prompts to elicit specific and reliable responses from LLMs, followed by post-processing of the generated outputs to derive structured and precise results. This process becomes particularly critical when LLMs are integrated as components of within more complex systems, such as for function calling or multi-agent collaboration, where inaccuracies in earlier stages can propagate through the workflow, adversely affecting overall system performance.

An emerging requirement for LLM-based applications is to support structured output. Although certain proprietary models~\citep[e.g., GPT-4o][]{hurst2024gpt4osystemcard} inherently support structured output through specialized training and constrained decoding (CD) mechanisms, replicating this capability for other LLMs presents significant challenges. In particular, for platforms that serve a diverse set LLMs, including open-weight and proprietary models, post-training each model for structured output without undermining general purpose performance is neither feasible nor scalable. This challenge is further exacerbated by the continual introduction of new models, making per-model adaptation impractical in the dynamic deployment environment.

\begin{figure}
    \centering
    \includegraphics[width=1\linewidth] {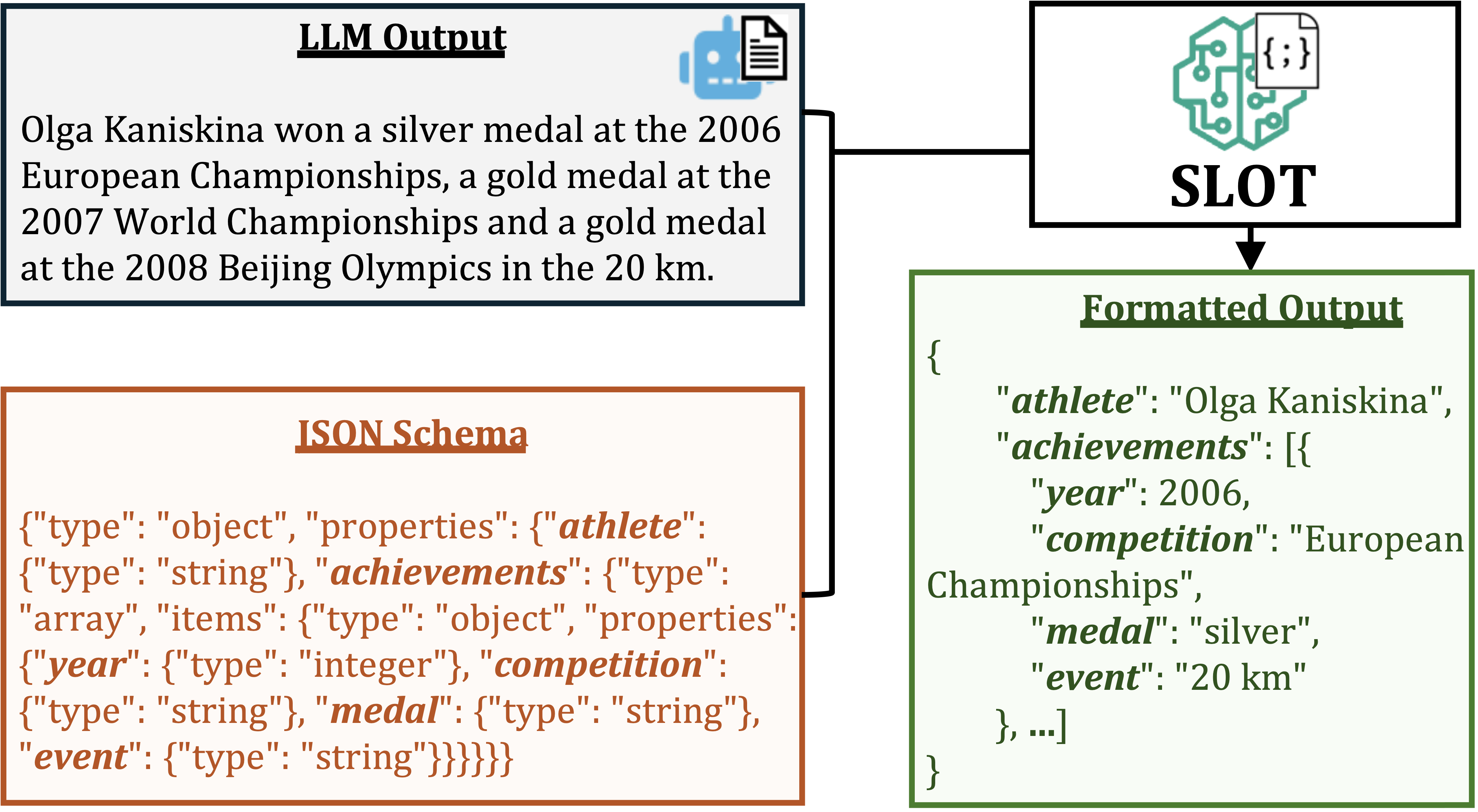} 
    \caption{\llmformatter~converts a textual LLM response into structured JSON with a pre-defined schema.}
    \label{fig:toc}
\end{figure}

To tackle these challenges, we introduce \llmformatter~for converting unstructured text into structured output. Different from existing approaches that relies on LLM post-training and constrained decoding,
\llmformatter~post-processes LLM's unstructured output by leveraging a lightweight language model, therefore being task and model-agnostic. 
Specifically, we fine-tune a lightweight language model to map unstructured text to the target schema, without modifying the underlying LLM. \llmformatter~ensures broad compatibility with both current and emerging models, regardless of their task specialization or output constraints, effectively bridging the gap between the generative flexibility of LLMs and the rigorous structural requirement for downstream integration in modern software systems.

Our main contributions are as follows:
\begin{itemize}[noitemsep,topsep=0pt,parsep=0pt,partopsep=0pt,leftmargin=*]

    \item We introduce \llmformatter~for structured output conversion applicable to any textual LLM outputs.
    \item We develop a synthetic data pipeline for diverse training data creation.
    \item We present evaluation metrics for \llmformatter~ covering schema accuracy and content similarity. 
    \item We demonstrate lightweight models including Llama-3.2 (1B/3B) and Mistral-7B outperform existing solutions across evaluation dimensions through supervised fine-tuning. 
\end{itemize}



\section{Background and Problem Formulation}
\label{sec:background}
\paragraph{Structured Output and JSON Schema.}
In the context of LLM, the term \emph{structured output} means that model-generated content conforms to a pre-defined, machine readable format rather than free-form natural language~\cite{dong2024xgrammarflexibleefficientstructured,Jiang2023StructGPTAG,openai2024introducingstructuredoutputs}. 
Generating structured data from unstructured inputs enables the ability of LLM-based applications to answer questions via function calling, extract structured data and build multi-step agent workflows that allow LLMs to take actions.
One of the most widely-used structure format is \emph{JSON Schema}, which is a declarative language to define structure and constraints for JSON data. JSON schema maintains a consistent pattern, making it easier to ensure data validity and exchange structured data between applications. In this paper, we focus specifically on JSON schema due to its wide adoption. 

\paragraph{Existing Approaches.}
To enable the model generation to conform to a target JSON schema, existing works mainly opted for three approaches, which have their strengths and weaknesses. Direct prompting, although straightforward, often yields lower performance~\cite{tam2024letspeakfreelystudy}. Constrained decoding methods, exemplified by works like~\cite{dong2024xgrammarflexibleefficientstructured,openai2024introducingstructuredoutputs}, does not rely on dedicated post-training, but only supports limited types of JSON schemas and may underperform training-based methods. Post-training generally improves performance compared to constrained decoding, but also requires access to model weights. The training itself adds complexity, and is not scalable for adaptation to new LLMs.\footnote{We should note that training-based methods can be combined with constrained decoding methods, which leads to improved performance as we observed in experiments (\cref{sec:experiments}).}

\paragraph{Problem Formulation.}
We focus on the post-processing setting (\cref{fig:toc}), where we assume access to an LLM's unstructured output.
We define our task as a text-to-structure problem, i.e., \emph{to transform an existing LLM's free-form text output into a structured format according to a specified JSON schema}. A more detailed comparison between \llmformatter~and existing mechanisms is presented in Appx (\cref{fig:method_comparison_appendix}). 

Given $x$ as an input text and $f$ as the formatting specification, let $M_{\theta}$ be a generative model parameterized by $\theta$: $M_{\theta}(x, f) \rightarrow y'$, and we seek to optimize the probability distribution $P\big(y | x, f; \theta)$ where $y$ is the groundtruth structured output and $y'$ the model's structured output. Note that the input text $x$ typically represents a response from an LLM rather than a user query.
Our post-processing based method is flexible and does not require access to LLM model weights, making it suitable for the case of serving a diverse array of LLMs in a platform. 

\section{Evaluation Framework}
\label{sec:evaluation_metrics}
The evaluation of structured output is inherently multi-faceted, as structure-wise, the output needs to adhere to the target JSON schema; while also not derailing from the unstructured input's semantic meaning. Therefore, we design our evaluation framework with the focus on (1) \emph{schema accuracy} and (2) \emph{content similarity}.

\paragraph{Schema Accuracy.}
We define schema accuracy as $A_s(y' \mid f)$: whether the response JSON string $y'$ exactly matches the user-demanded schema $f$ in terms of key strings and value types. To accurately assess the LLM's formatting capability, we directly evaluate response $y'$. The response $y'$ must be a valid JSON string by itself to be considered correct.

\paragraph{Content Similarity.}
The desired structured output should not derail from the unstructured input in terms of semantic meaning.
Assume that the groundtruth JSON structured output $y$ is provided, we can compute the content similarity between groundtruth output $y$ and model's output $y'$ as $\textit{sim}_C(y, y')$. 
In practice, we evaluate the semantic similarity between $y$ and $y'$. 
For each value in the prediction $y'$, we compute its semantic similarity using a pre-trained Sentence-BERT model~\cite{reimers2019sentencebertsentenceembeddingsusing,Zhangetal2020BERTScore} against the corresponding value (with matching key; including all ancestor keys for nesting structures) in the groundtruth JSON $y$. Missing keys result in a score of 0. The average of these scores represents the ``soft-precision", denoted $\textit{sim}_P(y, y')$. Similarly, we calculate ``soft-recall", denoted $\textit{sim}_R(y, y')$, by averaging SBERT scores for gold JSON values against their counterparts in the prediction. The content similarity score is therefore the harmonic mean of soft-precision and soft-recall (details in \cref{appendix:content_similarity_details}):
\begin{equation*}
    \begin{aligned}
    sim_C(y, y') = 2\times\frac{\textit{sim}_P(y, y') \times \textit{sim}_R(y, y')}{\textit{sim}_P(y, y')+\textit{sim}_R(y, y')}
    \end{aligned}
\end{equation*}
Additional evaluation dimensions are considered in related work (see \cref{appendix:evaluation_details}). For instance, task performance metrics ~\cite{tam2024letspeakfreelystudy,beurerkellner2024guidingllmsrightway, jiang2024sketchtoolkitstreamliningllm, shorten2024structuredragjsonresponseformatting,geng2025jsonschemabenchrigorousbenchmarkstructured} assess impact on original tasks but may lack generalizability. Computational efficiency metrics, such as latency or speed-up ~\cite{willard2023efficientguidedgenerationlarge, geng2025jsonschemabenchrigorousbenchmarkstructured}, prioritize algorithmic performance over output quality. Furthermore, JSON validity checks ~\cite{zhou2023instructionfollowingevaluationlargelanguage,beurerkellner2024guidingllmsrightway,jiang2024sketchtoolkitstreamliningllm,agarwal2025thinkinsidejsonreinforcement, geng2025jsonschemabenchrigorousbenchmarkstructured, ijgi13110405,xia2024fofobenchmarkevaluatellms} ensure syntax but often do not guarantee semantic correctness. These metrics alone may not simultaneously address the dual requirements of structural correctness and semantic preservation to structured output evaluation, hence are not focused in this work. 

\section{Data Pipeline}
\label{sec:dataset_creation}

\begin{figure}[t]
    \centering 
    \begin{minipage}[t]{0.48\textwidth}
        \vspace{0pt} 
        \footnotesize
        \centering
        \rowcolors{2}{gray!15}{white}
        \begin{tabular}{l|cccc}
            \hline
            \textbf{Dataset}  &\textbf{Train} & \textbf{Val} &\textbf{Test}\\
            \hline
            WebNLG  & 13,211 & - & 2,779\\
            E2E NLG  & 12,568 & - & 2,347\\
            WikiBio  & 25,000 & 650 & 2,500\\
            ToTTo $^{\diamond}$  & - & - & 500\\
            HF GitHub Issues $^{\diamond}$ & - & - & 1,000\\
            Synthetic $^{\star}$  & 126,000 & - & -\\
            \hline
            Total&176,779&650&9,126 \\
            \hline
        \end{tabular}
        \label{tab:data}
    \end{minipage}
    \hfill
    \begin{minipage}[t]{0.33\textwidth}
        \vspace{0pt}  
        \centering
        \includegraphics[width=\textwidth]{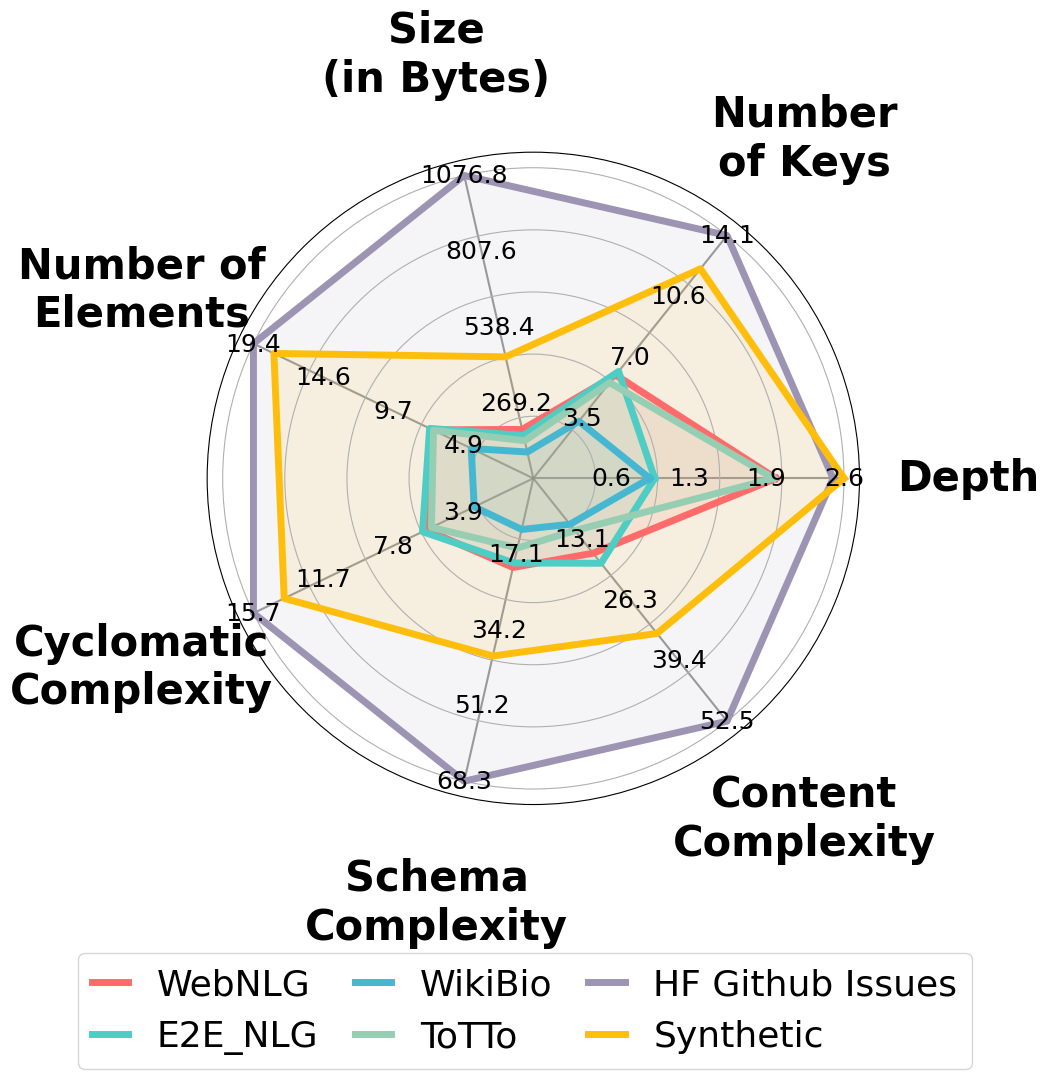}
    \end{minipage}
    \caption{Top: Statistics of the datasets used in experiments. $^{\diamond}$ refers to partially synthesized for repurposing and $^{\star}$ refers to fully synthesized. Bottom: JSON complexity in different dimensions. }
    \label{fig:data_complexity}
\end{figure}

To enable training of \llmformatter, we designed a data pipeline (Appx \cref{fig:data_pipeline_appendix}) to repurpose existing public datasets for structured output, and synthesize challenging datasets.
The main challenge in creating datasets for text-to-structure tasks lies in obtaining high-quality $(x,f,y)$ triples that: (1) cover diverse domains and text styles, (2) contain valid and well-formed structured outputs, and (3) ensure the structured outputs faithfully represent information present in the input text without hallucination. We address these challenges through a combination of \emph{synthetic data generation} and \emph{careful curation of public datasets}. 
Our final dataset consists of a synthetic training set mixture of 126K examples and multiple test sets totaling over 9K examples across diverse domains and formats. 
To analyze the quality of the synthetic data versus existing public datasets, we define seven dimensions to characterize the relative JSON complexity of different datasets, including depths, number of keys, number of elements, size (bytes), cyclomatic complexity, schema complexity and content complexity (details in \cref{sec:data_complexity_calculation_details}), and the breakdowns can be found from Figure \ref{fig:data_complexity}.

\paragraph{Test Benchmark Curation.}
\label{subsec: test_benchmark_curation}
To evaluate LLM Formatter comprehensively, we curated test sets from 5 public datasets spanning different domains and complexity levels: 
\begin{itemize}[noitemsep,topsep=0pt,parsep=0pt,partopsep=0pt,leftmargin=*]
\item \textbf{WebNLG} \citep{zhou-lampouras-2020-webnlg}: Originally designed for structured data-to-text generation, the dataset contains factual descriptions about entities and their relationships.
\item \textbf{E2E NLG} \citep{puzikov-gurevych-2018-e2e}: Contains restaurant domain descriptions paired with attribute-value structures.
\item \textbf{WikiBio} \citep{lebret-etal-2016-neural}: A dataset of biographical sentences paired with infobox-style structured data from Wikipedia.
\item \textbf{ToTTo} \citep{parikh2020totto}: A table-to-text dataset containing highlighted tables cells paired with corresponding descriptive sentences.

\item \textbf{HF GitHub Issues}: To evaluate model performance on complex, real-world scenarios, we curated a challenging benchmark from Hugging Face Transformers GitHub issues\footnote{\url{https://api.github.com/repos/huggingface/transformers/issues}}. This dataset captures diverse software issue elements (e.g., system/error information, reproduction steps, code snippets, expected behaviors). It exhibits significantly higher complexity in structure and technical content compared to existing benchmarks, providing a rigorous test of the model's ability to handle intricate nested structures and technical inputs.
\end{itemize}
We defer data curation details and specific examples to~\cref{appendix: dataset_curation_details_and_examples}.

\paragraph{Training Data Creation.}
\label{subsec: training_data_creation}

We use Claude-3.5-Sonnet to systematically generate synthetic training data, ensuring broad coverage by sampling across five variables related to diversity and quality: industry vertical, JSON complexity, text length style, genre, and text type (\cref{sec:appendix-training-data-details}). For each generation batch, we first sample 1-3 demonstrations from the WikiBio training set. We then randomly sample values for each control dimension to define target characteristics, incorporating these into our prompt (\cref{appendix:prompt:data_generation_valdiation}) to guide generation. The prompt encourages format adherence, content diversity, and semantic coherence between input text and structured output. Temperature sampling ($T\in[0,0.5]$) is applied to balance creativity and consistency.

To ensure data quality, we use a two-stage validation process to filter unfactual / hallucinated generations. Stage 1 checks format validity, ensuring $f$ (schema) and $y$ (output) are parsable, structurally consistent, and adhere to type agreements and required structural relationships. Because content validity is challenging for semantic text-to-structure tasks where rule-based/fuzzy matching fails, Stage 2 employs an LLM validator. This validator checks if each structured output field's content is reasonably inferable from the source text ($x$). This data synthesis pipeline yielded 126K high-quality training examples for training \llmformatter~(\cref{sec:experiments}).


\section{Training \llmformatter}
\label{sec:method}
\begin{figure}
    \centering
    \includegraphics[width=1\linewidth]{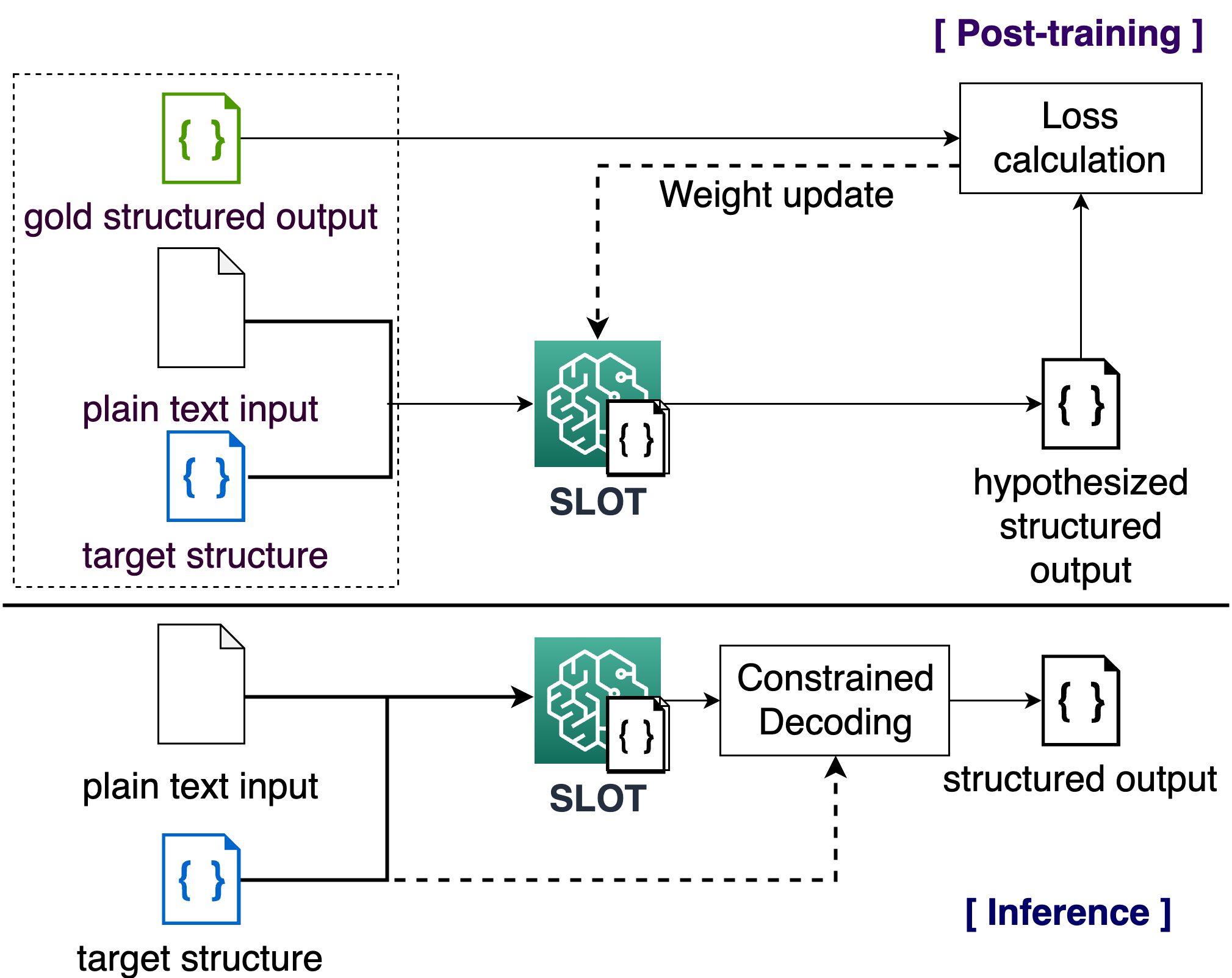}
    \caption{Our proposed framework for post-training and hosting an LLM for structured output generation.}
    \label{fig:framework}
\end{figure}

Our primiary goal is to train \llmformatter~to accurately transform unstructured text input into a structured representation conforming to a specified schema, optimizing for the evaluation metrics detailed in \cref{sec:evaluation_metrics}, which include structural validity, schema compliance, and semantic accuracy of extracted values. To achieve this, we employ Supervised Fine-tuning (SFT) as our core training strategy.

We build \llmformatter~upon a pre-trained decoder language model (e.g., Llama-3 and Mistral), which provides a strong foundation in natural language understanding and generation, significantly reducing the need for training from scratch and allowing us to focus computational resources on adapting the model to the specific nuances of the text-to-structure task. The autoregressive nature is well-suited for generating sequential structured outputs like JSON strings token by token.

We adopt SFT due to its effectiveness in teaching pre-trained models to follow instructions and generate outputs in specific formats when provided with high-quality input-output examples \citep{ouyang2022training}. Our task perfectly fits this paradigm: we have source text ($x_i$), a clear instruction embedded within the schema specification ($f_i$), and a desired target output ($y_i$).

The training dataset $\mathcal{D}=\{ (x_i, f_i, y_i)\}$, synthesized as described in \cref{subsec: training_data_creation}, forms the basis for SFT. During fine-tuning, each triple $(x_i, f_i, y_i)$ from dataset $\mathcal{D}$ is formatted into an instruction-following prompt. A typical format concatenates the task instruction, the target schema, and the input text into a single input sequence for the model.
We then apply the standard causal language modeling loss on the target structured output $y_i$ only, masking out the input prompt tokens from loss calculation.

Unlike prior works \citep[\interalia]{jiang2024sketchtoolkitstreamliningllm,agarwal2025thinkinsidejsonreinforcement}, our approach is task-agnostic -- \llmformatter\space decouples formatting from task-specific generation, which ensures compatibility with any LLM output for any task without modification, while avoiding generation quality degradation from constrained decoding. This positions \llmformatter~as a universal adapter between general-purpose LLMs and structured data applications.

\section{Experiments}
\label{sec:experiments}

\begin{table*}[h!]
    \centering
    \footnotesize
    \resizebox{\textwidth}{!}{
    \begin{tabular}{lrrrrrr|rrrrrr}
        \toprule
        \, & 
        \multicolumn{6}{c|}{\textbf{Schema Accuracy (\%)}} & 
        \multicolumn{6}{c}{\textbf{Content Similarity (\%)}} \\
        \cmidrule{2-13}
\shortstack[l]{LLM\\+Settings} 
& \shortstack{Web-\\NLG} 
& \shortstack{E2E-\\NLG} 
& \shortstack{Wiki-\\Bio} 
& \shortstack{ToTTo} 
& \shortstack{GitHub\\Issues} 
& \textbf{Avg} 
& \shortstack{Web-\\NLG} 
& \shortstack{E2E-\\NLG} 
& \shortstack{Wiki-\\Bio} 
& \shortstack{ToTTo} 
& \shortstack{GitHub\\Issues} 
& \textbf{Avg} \\

\rowcolor{gray!40}
\multicolumn{13}{l}{LLM Baselines}\\
\rowcolor{gray!15}
\multicolumn{13}{l}{Claude-3.5-Haiku}\\
+Prompting & 50.3&98.9&99.3&99.0&97.6&89.0 &40.5&90.4&86.8&96.4&89.4&80.7\\


\rowcolor{gray!15}
\multicolumn{13}{l}{Claude-3.5-Sonnet}\\
+Prompting&11.8&77.0&97.8&91.7&95.3&74.7    &9.4&92.8&85.4&91.1&90.9&73.9\\


\rowcolor{gray!15}
\multicolumn{13}{l}{Qwen-2.5-14B}\\
+Prompting&0.0&0.0&0.0&0.0&0.1&0.0&0.0&0.0&0.0&0.0&0.1&0.0\\
+OT&100&100&100&99.8&76.7&95.3&79.3&89.8&86.7&96.8&70.4&84.6\\
+XG&100&100&100&99.4&97.1&99.4&53.5&86.6&86.7&96.8&89.4&82.6\\

\rowcolor{gray!15}
\multicolumn{13}{l}{Qwen-2.5-32B}\\
+Prompting&0.0&0.0&0.0&0.0&0.5&0.1&0.0&0.0&0.0&0.0&0.5&0.1\\
+OT&100&100&100&99.6&76.4&95.2&80.7&93.9&91.1&97.6&70.2&86.7\\
+XG&100&100&100&99.4&97.1&99.3&80.6&93.8&91.0&97.5&89.6&90.5\\

\rowcolor{gray!15}
\multicolumn{13}{l}{Llama-3.3-70B}\\
+Prompting&3.2&54.1&32.6&28.8&0.9&23.9&2.6&48.5&30.7&28.4&0.9&22.2\\
+OT&100&100&100&99.0&78.0&95.4&80.2&85.3&90.6&97.4&70.3&84.8\\
+XG&100&100&100&98.8&99.1&\textbf{99.6}&80.3&85.7&91.0&97.4&91.0&89.1\\


\rowcolor{gray!40}
\multicolumn{13}{l}{\llmformatter}\\
\rowcolor{gray!15}
\multicolumn{13}{l}{Llama-3.2-1B}\\
+Prompting&0.0&0.0&42.2&2.1&3.5&9.6    &0.1&0&32.0&1.5&1.7&7.1\\

\cellcolor{lightCornflowerBlue}{+SFT}&97.7&100&99.4&95.4&52.2&88.9    &79.8&98.2&94.3&90.0&46.2&81.7\\

+OT&100&100&99.9&97.9&56.9&90.9    &72.2&76.1&85.6&93.7&51.3&75.8\\

\cellcolor{lightCornflowerBlue}{+SFT \& OT} &100&99.5&100&100&15.4&83.0 &83.3&94.6&93.9&95.7&13.4&76.2\\

+XG &100&100&99.8&91.7&78.1&93.9    &68.6&76.3&83.7&90.3&73.4&78.4\\

\cellcolor{lightCornflowerBlue}{+SFT \& XG}&100&100&100&100&81.2&96.2 &82.3&98.2&94.3&95.3&78.1&89.6\\

\rowcolor{gray!15}
\multicolumn{13}{l}{Llama-3.2-3B}\\
+Prompting&0.0&0.0&1.5&0.0&0.1&0.3    &0.0&0.0&0.0&0.0&0.0&0.0\\

\cellcolor{lightCornflowerBlue}{+SFT}&99.8&99.7&100&98.6&75.6&94.7    &85.3&98.1&95.9&94.7&69.4&88.7\\

+OT&100&100&100&96.5&56.9&90.7     &74.3&76.1&87.4&95.3&56.3&77.9\\

\cellcolor{lightCornflowerBlue}{+SFT \& OT}&99.3&99.8&99.4&99.0&36.9&86.9  &84.9&96.0&94.4&95.5&34.7&81.1\\

+XG &100&100&99.6&98.3&84.0&96.4    &67.0&77.5&84.0&94.4&77.7&80.1\\

\cellcolor{lightCornflowerBlue}{+SFT \& XG}&100&100&100&99.8&91.5&\underline{98.2}  &85.5&98.1&95.6&96.7&86.1&92.4\\

\rowcolor{gray!15}
\multicolumn{13}{l}{Mistral-7B-v0.2} \\ 
+Prompting&0.0&0.0&0.0&0.0&0.0&0.0  &0.0&0.0&0.0&0.0&0.0&0.0\\

\cellcolor{lightCornflowerBlue}{+SFT}&99.5&100&100&98.3&93.1&\underline{98.2}      &87.0&98.7&96.2&96.7&85.9&\underline{92.9}\\

+OT&100&100&99.9&88.6&60.7&89.8 &73.8&83.0&86.3&87.6&58.4&77.8\\

\cellcolor{lightCornflowerBlue}{+SFT \& OT}&100&100&99.9&97.9&70.9&93.7    &86.6&94.9&94.7&96.0&64.7&87.4\\

+XG&100&100&99.8&88.2&73.1&92.2 &74.0&82.9&85.9&87.9&70.2&80.2\\

\cellcolor{lightCornflowerBlue}{+SFT \& XG} &100&100&100&99.8&97.5&\textbf{99.5}    &87.4&98.7&96.3&98.0&89.7&\textbf{94.0}\\
   \hline
    \end{tabular}
    }
    \caption{Average percentage schema accuracy and content similarity of different base and fine-tuned LLMs. OT denotes Outlines, XG denotes XGrammar.}
    \label{tab:sec-experiment}
\end{table*}
In this section, we present a comprehensive evaluation of \llmformatter. We first introduce our experiment setup (\cref{subsec:experiment_setting}), followed by detailed result analysis and discussion (\cref{subsec:experiment_result}).
\subsection{Experiment Setting}
\label{subsec:experiment_setting}

\paragraph{Methods Compared.}
We evaluate \llmformatter~against state-of-the-art methods for structured output generation, which fall into two categories. First, we use proprietary LLMs, specifically Claude-3.5-Haiku and Claude-3.5-Sonnet~\cite{claude35haiku,claude35sonnet} with standard prompting to establish strong baselines. Second, we experiment with open-weight LLMs with constrained decoding, including Qwen-2.5 (14B, 32B)~\cite{qwen2025qwen25technicalreport} and Llama-3.3-70B~\cite{grattafiori2024llama3technicalreport} using two leading constrained decoding frameworks: Outlines~\cite{willard2023efficientguidedgenerationlarge}\footnote{\url{https://github.com/dottxt-ai/outlines}} and XGrammar~\cite{dong2024xgrammarflexibleefficientstructured}\footnote{\url{https://github.com/mlc-ai/xgrammar}}. We also include direct prompting results for reference in \cref{appendix:prompt:direct_prompting}.

We employ three lightweight language models in their \texttt{Instruct} versions:  Llama-3.2 (1B / 3B)~\cite{grattafiori2024llama3technicalreport}, and Mistral-7B-v0.2~\cite{jiang2023mistral7btechnicalreport}. We also evaluate \llmformatter~ when combined with constrained decoding methods in the inference time, which provides an opportunity to understand how model-based and rule-based approaches can complement each other.

\paragraph{Training Setup.}
For \llmformatter~training, we employ LoRA~\cite{hu2022lora} fine-tuning with a standard language modeling objective on completions. We selected checkpoints based on a balanced metric combining schema accuracy and content similarity, evaluated on a validation subset of 650 instances from WikiBio. For inference, We used vLLM \cite{kwon2023efficient} to optimize generation throughput. The complete hyperparameters for both training and inference are detailed in Appx \cref{tab:hyperparams}.

\subsection{Results and Analysis}
\label{subsec:experiment_result}
Table \ref{tab:sec-experiment} presents our main results, with additional details provided in \cref{tab:schema_acc_appendix} and \cref{tab:content_sim_appendix}. Our analysis reveals several key findings:

\paragraph{Schema Accuracy and Content Similarity.}

Proprietary models establish competitive baseline performance, with Claude-3.5-Sonnet achieving 74.7\% schema accuracy and 73.9\% content similarity on average. However, even these advanced models fall short of the reliability required for critical applications like function calling or multi-agent collaboration, suggesting potential limitations in pure prompting approaches. In contrast, small open-weight models perform extremely poorly \texttt{via} direct prompting -- Llama-3.2-1B, 3B, and Mistral-7B-v0.2 achieve near-zero schema accuracy and content similarity across most datasets.

\llmformatter~dramatically transforms this landscape through supervised fine-tuning. The 1B parameter model achieves 88.9\% overall schema accuracy (comparable to Claude-3.5-Haiku) and 81.7\% content similarity (exceeding Claude-3.5-Sonnet's 73.9\%). More impressively, Mistral-7B-v0.2 reaches 98.2\% schema accuracy and 92.9\% content similarity, surpassing all proprietary baselines on both metrics. These results demonstrate that specialized training can outperform general capabilities of much larger models for structured generation tasks. Dataset complexity significantly impacts performance across all methods. The GitHub Issues dataset appears to be the most challenging across all configurations, given the prevalence of deeply nested structures and technical content (\cref{fig:data_complexity}). Yet even here \llmformatter~ shows remarkable gains -- Mistral-7B improves from 0\% to 93.1\% schema accuracy. Conversely, E2E NLG consistenly yields the highest content similarity across models, likely due to its simpler descriptions with well-defined attributes. 

\paragraph{Synergy Between Fine-Tuning and Constrained Decoding.} We observe that constrained decoding methods reveal interesting algorithmic tradeoffs between structure and semantics. XGrammar performs markedly better on complex nested structures (achieving 97.1\% schema accuracy on GitHub issues with Qwen-2.5-32B compared to Outlines' 76.4\%). This advantage stems from XGrammar's adaptive token caching strategy that efficiently handles context-independent tokens and context expansion. The byte-level pushdown automaton approach with persistent execution stack enables efficient validation of complex structures, while still allowing for semantic coherence during generation. Outlines offers efficiency advantage through finite-state indexing for simpler structures, but its traditional pushdown automaton faces scalability challenges with deeply nested structures -- resulting longer processing times and timeout errors with complex schemas. 

Most significantly, combining \llmformatter~ with constrained decoding creates a powerful synergy. Mistral-7B-v0.2 with SFT + XGrammar achieves near-perfect results: 99.5\% schema accuracy and 94.0\% content similarity. Even the 1B parameter model reaches impressive performance (96.2\% schema accuracy, 89.6\% content similarity) with this combined approach, demonstrating that structured output generation does not necessarily require massive model sizes, but rather targeted training and appropriate constraints. This synergy occurs because SFT teaches the model to inherently produce well-structured and semantically appropriate outputs, while constrained decoding provides a complementary guarantee of structural validity, which is particularly pronounced in smaller models. Detailed error analysis is in \cref{appendix:error_analysis}.

\paragraph{Impact of Training Data.}
Our detailed analysis of training data contributions (\cref{tab:schema_acc_appendix}, \cref{tab:content_sim_appendix}) reveals that the diversity and complexity of synthetic data plays a crucial role in \llmformatter's success. For instance, with Llama-3.2-1B, public data alone achieves 63.1\% schema accuracy and 68.3\% content similarity. Synthetic data alone performs substantially better for schema accuracy, reaching 89.6\%, while achieving 74.4\% content similarity. The combined dataset demonstrates complementary benefits, achieving 89.0\% schema accuracy and 81.7\% content similarity, with particular gains in content preservation. This pattern holds across model sizes, with the strongest results typically coming from the combined dataset.

\section{Related Works}
\label{sec:related}
The transition from natural language output to structured formats presents significant challenges that researchers have approached through various methods within the following categories:

\paragraph{Direct Prompting.} The most straightforward approach involves explicitly instructing LLMs to generate structured outputs, supported by frameworks like LangChain \cite{Chase_LangChain_2022}, LlamaIndex \cite{Liu_LlamaIndex_2022}, and Instructor \cite{Liu_instructor_2022}. However, \citet{tam2024letspeakfreelystudy} demonstrated that this approach often produces invalid structures and can lower performance on reasoning-intensive tasks due to the additional cognitive load of format adherence.

\paragraph{Constrained Decoding.} To guarantee structural validity, constrained decoding techniques \citep{beurerkellner2024guidingllmsrightway,Liu_2024} guide the generation process using formal grammars. Tools like llama.cpp \cite{GerganovLiu_llamacpp_2023}, Guidance \cite{Guidance_AI_2023}, XGrammar \cite{dong2024xgrammarflexibleefficientstructured}, and Outlines \cite{willard2023efficientguidedgenerationlarge} implement this approach. While effective at ensuring valid outputs, these methods typically increase inference latency and may impact task performance due to the restrictive nature of the constraints.

\paragraph{Post-Training Approaches.} Task-specific fine-tuning has shown success in structured output generation, particularly for tasks like named-entity recognition \cite{wang2023gptnernamedentityrecognition} and information extraction \cite{wang2023instructuiemultitaskinstructiontuning}. However, this approach requires separate training efforts for each task-format combination, limiting its scalability.

\paragraph{Hybrid Solutions.} Recent work has explored combining multiple techniques to balance reliability and performance. \citet{jiang2024sketchtoolkitstreamliningllm} demonstrated success in fine-tuning with diverse schema datasets, while \citet{agarwal2025thinkinsidejsonreinforcement} combined reinforcement learning with supervised fine-tuning for schema-constrained tasks.

As illustrated in Appx ~\cref{fig:method_comparison_appendix}, these existing approaches are typically tied to specific LLMs, requiring reimplementation for each new model. Additionally, previously launched LLMs cannot benefit from these improvements without significant modification. Our proposed \llmformatter~addresses these limitations by decoupling output formatting from the natural language task, offering a model-agnostic solution that maintains task performance while ensuring structural validity.

\section{Conclusion}
\label{sec:conclusion}
We introduced \llmformatter, a model-agnostic and task-agnostic framework for generating structured outputs from LLMs. While our solution can be theoretically applied to any specified schema, we focused on JSON as our primary use case and developed two objective metrics -- Schema Accuracy and Content Similarity -- to evaluate the quality of generated JSON outputs. Through a comprehensive data pipeline, we synthesized and validated training data spanning diverse text lengths, styles, industry verticals, and JSON complexity levels. Our experimental results demonstrate that with supervised fine-tuning, lightweight open-weight language models can outperform larger proprietary models and exhibit strong generalizability. This finding underscores the efficacy of training-based approaches for structured output generation, potentially improving the accessibility of structured data in real-world applications.
Future work will focus on investigating more sophisticated data synthesis strategies and exploring advanced post-training methodologies, including preference tuning and reinforcement learning.

\section*{Limitations}
\label{sec:limitation}
Due to licensing agreements with proprietary LLM providers, we are unable to publish our synthetic training and test datasets although we provide the prompts used for data synthesis and validation. Furthermore, our data synthesis approach primarily focused on cases where the input schema and gold structured output capture the majority of entities and relationships present in the input text. We did not extensively explore scenarios where only a subset of the textual information needs to be structured, or where complex filtering or transformation of the input content is required. 

Finally, we note that an ideal evaluation of \llmformatter~would involve end-to-end testing, where the input text is generated by a preceding LLM instead of using arbitrary texts that emulate LLM output. These limitations suggest potential areas for future work in handling more diverse LLM output and structuring requirements.

\bibliography{custom}

\appendix
\label{sec:appendix}

\onecolumn

\section{Related Work Details}
\label{appendix:related_work_details}
\begin{figure}[h!]
    \centering
    \includegraphics[width=\linewidth]{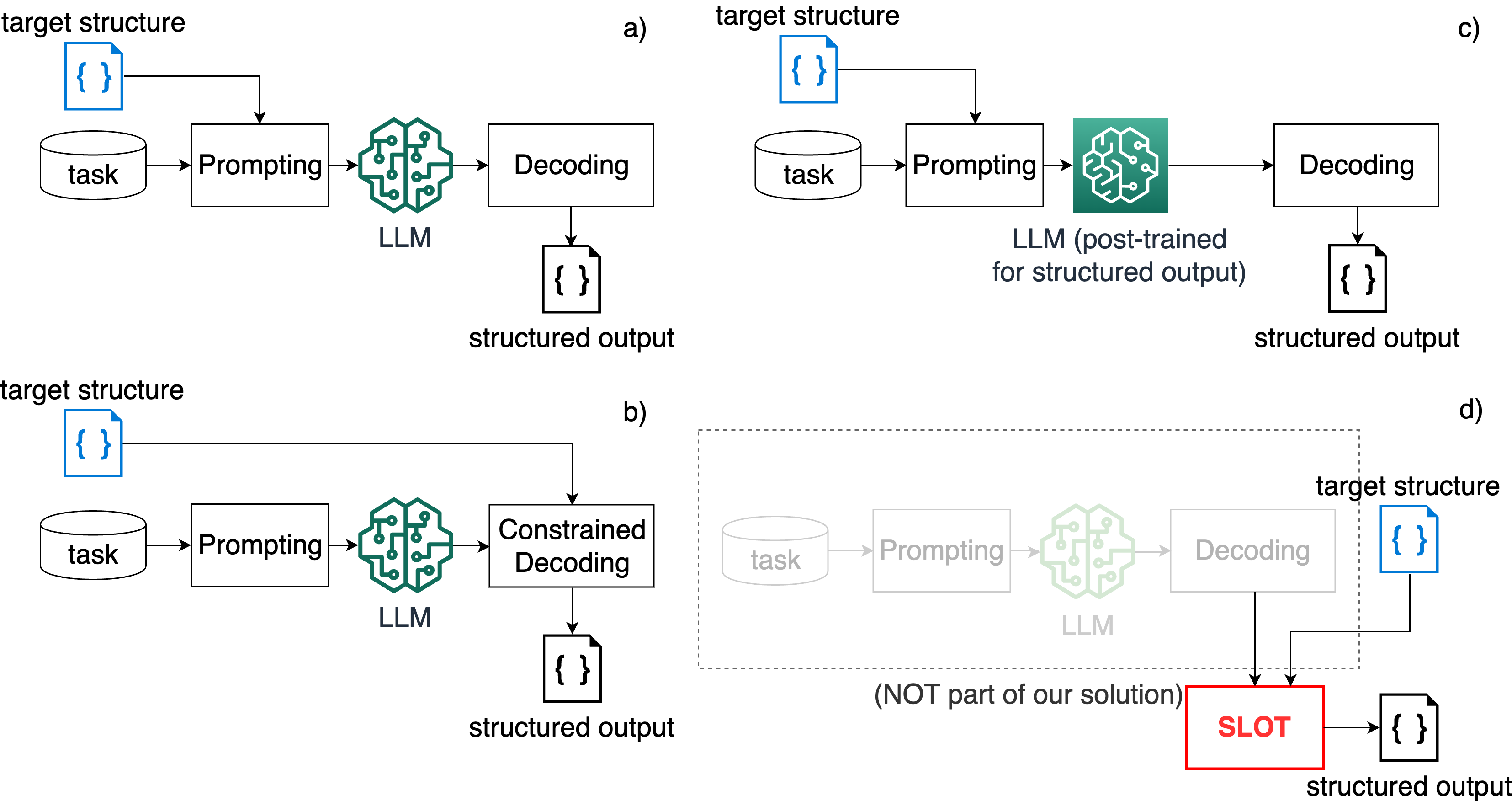}
    \caption{Approaches for LLM structured outputs. a) prompting LLM form structured output, b) constrained decoding, c) post-training, and d) \llmformatter.}
    \label{fig:method_comparison_appendix}
\end{figure}

\section{Detailed Experiment Results}
\label{appendix:experiment_results}
\begin{table*}[h!]
    \centering
    \footnotesize
    \rowcolors{2}{gray!15}{white}
    \resizebox{1\textwidth}{!}{
        \begin{tabular}{ p{3.75cm}|p{2.5cm}||c|c|c|c|c||c }
            \hline
           \multirow{2}{3.75cm}{\textbf{LLM \& Settings} \newline}&\multirow{2}{2.5cm}{\textbf{SFT dataset} \newline}&\multicolumn{6}{c}{\textbf{Schema Accuracy (\%)}} \\
           \cline{3-8}
           && WebNLG&E2E NLG&WikiBio&ToTTo&GitHub&Avg\\
           \hline




           Llama 3.2 11B&-&0.47&0.13&45.88&60.58&15.18&24.45\\



           Llama 3.2 1B& -&0&0&42.20&2.07&3.52&9.56\\

           &Public&99.28 & 91.86 & 99.44 & 13.49 & 11.26 & 63.07 \\

           &Synthetic& 99.42	&99.96	&98.4	&95.23	&54.87	&89.58 \\

           &Public \& Synthetic& 97.73	& 100 & 99.44 & 95.44 & 52.16 & 88.95 \\

           Llama 3.2 1B$_\text{ + Outlines}$&-&100&100&99.88&97.93&56.88&90.94\\
  
           &Public \& Synthetic& 100 & 99.53 & 99.96 & 100 & 15.38	&82.97 \\

           Llama 3.2 1B$_\text{ + XGrammar}$&-&100&100&99.8&91.7&78.09&93.92\\

           &Public \& Synthetic&100&100&100&100&81.21&96.24\\


           Llama 3.2 3B&-&0&0&1.48&0&0.10&0.32\\

           &Public& 99.46	&100	&99.88	&28.22	&34.07	&72.33 \\

           &Synthetic&100 & 99.79 & 98.92 & 94.81 & 82.31 & 95.17 \\

           &Public \& Synthetic&99.75 & 99.7 & 99.96 & 98.55 & 75.58 & 94.71 \\
           
           Llama 3.2 3B$_\text{ + Outlines}$&-&100&100&100&96.47&56.88&90.67\\

           &Public \& Synthetic&99.28 & 99.83 & 99.44 & 98.96 & 36.88 & 86.88 \\

           Llama 3.2 3B$_\text{ + XGrammar}$&-&100&100&99.6&98.34&84.02&96.39\\

           &Public \& Synthetic&100&100&99.96&99.79&91.46&98.24\\


           Mistral 7B (v0.2)&-&0&0&0&0&0&0\\

           &Public& 100	&100	&99.6	&11.83	&20.7	&66.43 \\

           &Synthetic& 99.03	& 99.96	& 97.88	& 98.76	& 88.14	& 96.75 \\

           &Public \& Synthetic& 99.5	& 100 & 99.96 & 98.34 & 93.07 & 98.17 \\

           Mistral 7B (v0.2)$_\text{ + Outlines}$&-&100&100&99.92&88.59&60.70&89.84\\

           &Public \& Synthetic& 100 & 100 & 99.92 & 97.93 & 70.85 & 93.74 \\

           Mistral 7B (v0.2)$_\text{ + XGrammar}$&-&100&100&99.8&88.17&73.07&92.21\\

           &Public \& Synthetic&100&100&100&99.79&97.49&99.46\\
           \hline
        \end{tabular}
    }
    \caption{Schema accuracy of different base and fine-tuned LLMs}
    \label{tab:schema_acc_appendix}
\end{table*}

\begin{table*}[h!]
    \centering
    \footnotesize
    \rowcolors{2}{gray!15}{white}
        \begin{tabular}{ p{3.75cm}|p{2.5cm}||c|c|c|c|c||c }
            \hline
           \multirow{2}{3.75cm}{\textbf{LLM \& Settings} \newline}&\multirow{2}{2.5cm}{\textbf{SFT dataset} \newline}&\multicolumn{6}{c}{\textbf{Content Similarity (\%)}} \\
           \cline{3-8}
           && WebNLG&E2E NLG&WikiBio&ToTTo&GitHub&Avg\\
           \hline




       Llama 3.2 11B&-&0.42&0.20&35.63&59.30&9.90&21.09\\



       Llama 3.2 1B&-&0.08&0&32.03&1.54&1.69&7.07\\

       &Public& 80.59	& 89.91	& 94.35	& 51.13	& 25.48	& 68.29\\

       &Synthetic&74.46 & 75.51 & 82.78	& 90.68	& 48.77	 & 74.44 \\

       &Public + Synthetic&79.84 & 98.15 & 94.28 & 89.95 & 46.24 & 81.69 \\

       Llama 3.2 1B$_\text{ + Outlines}$&-&72.17&76.13&85.61&93.68&51.32&75.78\\

       &Public + Synthetic&83.34	& 94.62	& 93.87	& 95.73	& 13.43	& 76.20 \\

       Llama 3.2 1B$_\text{ + XGrammar}$&-&68.58&76.29&83.67&90.25&73.38&78.43\\

       &Public + Synthetic&82.32&98.17&94.31&95.32&78.09&89.64\\


       Llama 3.2 3B&-&0&0&0.02&0&0&0\\

       &Public& 84.87 & 98.3 & 95.5 & 79.48 & 50.99 & 81.83 \\

       &Synthetic&78.42	& 81.51	& 85.56	& 91.35	& 74.86	& 82.34 \\

       &Public + Synthetic& 85.32 & 98.08 & 95.88 & 94.73 & 69.43 & 88.69 \\

        Llama 3.2 3B$_\text{ + Outlines}$&-&74.3&76.06&87.36&95.26&56.25&77.85\\

       &Public + Synthetic&84.93 & 96.01	& 94.4	& 95.45	& 34.67	& 81.09 \\

        Llama 3.2 3B$_\text{ + XGrammar}$&-&67&77.49&84.02&94.42&77.65&80.12\\

       &Public + Synthetic&85.53&98.06&95.63&96.73&86.14&92.42\\


       Mistral 7B (v0.2)&-&0&0&0&0&0&0\\

       &Public&85	&97.79	&95.86	&37.25	&25.49	&68.28 \\

       &Synthetic&77.91	&85.4	&86	&96.45	&81.66	&85.48 \\

       &Public + Synthetic& 87 & 98.72 & 96.24 & 96.73 & 85.94 & 92.93 \\
       
       Mistral 7B (v0.2)$_\text{ + Outlines}$&-&73.76&82.96&86.28&87.61&58.44&77.81\\

       &Public + Synthetic&86.6 & 94.88 & 94.69 & 96 & 64.66 & 87.37 \\

       Mistral 7B (v0.2)$_\text{ + XGrammar}$&-&73.95&82.93&85.93&87.92&70.16&80.18\\

       &Public + Synthetic&87.42&98.73&96.25&98.01&89.71&94.02\\
       \hline
    \end{tabular}
    \caption{Content similarity of different base and fine-tuned LLMs}
    \label{tab:content_sim_appendix}
\end{table*}

\newpage

\section{Evaluation Details}
\label{appendix:evaluation_details}
\label{sec:appendix-evaluation-details}

\subsection{Existing evaluation metrics for JSON structured outputs from LLMs}

\paragraph{Task performance} This approach treats structured output as an additional requirement alongside the original task (e.g., reasoning or information extraction). Researchers measure the impact of this requirement by comparing performance metrics (such as answer accuracy) with and without structured output constraints, or across different required formats. However, this evaluation method is task-specific and measures the LLM's ability to provide structured outputs for particular tasks rather than its general formatting capability (e.g., in \citet{tam2024letspeakfreelystudy}, \citet{beurerkellner2024guidingllmsrightway}, \citet{jiang2024sketchtoolkitstreamliningllm}, \citet{shorten2024structuredragjsonresponseformatting}, \citet{geng2025jsonschemabenchrigorousbenchmarkstructured} etc.).
\paragraph{Latency or speed-up} Structured output requirements demand additional processing during next-token generation and/or constrained decoding. Works focusing on algorithmic efficiency often measure either latency (compared to baseline without structured output requirements) or speed-up (compared to other structured output tools) (e.g., in \citet{willard2023efficientguidedgenerationlarge}, \citet{geng2025jsonschemabenchrigorousbenchmarkstructured} etc.).
\paragraph{JSON validity} Many studies evaluate whether the LLM's responses are valid JSON, e.g. \citet{zhou2023instructionfollowingevaluationlargelanguage}, \citet{beurerkellner2024guidingllmsrightway}, \citet{jiang2024sketchtoolkitstreamliningllm}, \citet{agarwal2025thinkinsidejsonreinforcement}. Some examine schema compliance (e.g. \citet{geng2025jsonschemabenchrigorousbenchmarkstructured}) or value accuracy using exact match (\citet{agarwal2025thinkinsidejsonreinforcement}) or edit distance (\citet{ijgi13110405}). While some benchmarks like \citet{xia2024fofobenchmarkevaluatellms} relied on LLM-as-a-Judge (LLMaaJ) to evaluate the generated structured output's format and/or content validity, this approach's non-deterministic nature makes it unsuitable for industrial applications involving LLM agents or function calling.

\subsection{Details on content similarity metric}
\label{appendix:content_similarity_details}

The content similarity score in this work is defined as the harmonic mean of soft-precision and soft-recall:
\begin{equation*}
    \begin{aligned}
    sim_C(y, y') = 2\times\frac{\textit{sim}_P(y, y')\cdot \textit{sim}_R(y, y')}{\textit{sim}_P(y, y')+\textit{sim}_R(y, y')}
    \end{aligned}
\end{equation*}
There are several rationales behind the design choices of our content similarity. First, if we simply concatenate all the values into one large string, not only the order of concatenation is not trivial, such metric could be dominated by values of very long strings. Consider:

\begin{tcolorbox}[
    enhanced,
    colframe=blue!75!black,
    colback=yellow!5!white,
    drop shadow={opacity=0.3, xshift=1pt, yshift=-1pt},
    rounded corners=all,
    arc=1.2mm,
    left=1.5mm,
    right=1.5mm,
    top=1.5mm,
    bottom=0mm,
    boxsep=1mm,
    boxrule=0.5pt,
]
\begin{minipage}[t]{0.48\textwidth}
\textbf{\textcolor{blue!75!black}{\small{Gold:}}}
\begin{lstlisting}[
    basicstyle=\ttfamily\footnotesize, 
    backgroundcolor=\color{blue!5!white},
    breaklines=true, 
    columns=fullflexible, 
    frame=single,
    framesep=1mm,
    framerule=0.5pt,
    rulecolor=\color{gray!50},
    keepspaces=true,
]
{
  "error": "ValueError: This is a sample error message",
  "traceback": "Traceback (most recent call last): ......"
  // a very long error message 
}
\end{lstlisting}
\end{minipage}%
\hfill
\begin{minipage}[t]{0.48\textwidth}
\textbf{\textcolor{red!60!black}{\small{Prediction:}}}
\begin{lstlisting}[
    basicstyle=\ttfamily\footnotesize, 
    backgroundcolor=\color{red!5!white},
    breaklines=true, 
    columns=fullflexible, 
    frame=single,
    framesep=1mm,
    framerule=0.5pt,
    rulecolor=\color{gray!50},
    keepspaces=true,
]
{
  "error": "ValueError: This is a sample error message",
  "traceback": ""
}
\end{lstlisting}
\end{minipage}
\end{tcolorbox}
\noindent
In contrary, the pairwise SBERT works even when the numbers of entities in ground truth and prediction are different, and also weighs each value uniformly disregarding its length.

Second, requiring exact key matches, rather than using edit distances as in \citet{ijgi13110405}, helps identify cases where values are incorrectly swapped between similar keys. Here is one example we observed from the E2E NLG dataset:
\begin{tcolorbox}[
    enhanced,
    colframe=blue!75!black,
    colback=yellow!5!white,
    drop shadow={opacity=0.3, xshift=1pt, yshift=-1pt},
    rounded corners=all,
    arc=1.2mm,
    left=1.5mm,
    right=1.5mm,
    top=1.5mm,
    bottom=0mm,
    boxsep=1mm,
    boxrule=0.5pt,
]

\textbf{\textcolor{gray!60!black}{\small{Text:}}}
\begin{lstlisting}[
    basicstyle=\ttfamily\footnotesize, 
    backgroundcolor=\color{gray!5},
    breaklines=true, 
    columns=fullflexible, 
    frame=single,
    framesep=1mm,
    framerule=0.5pt,
    rulecolor=\color{gray!50},
    keepspaces=true,
]
"Near The Rice Boat you can visit coffee shop called Giraffe."
\end{lstlisting}

\tcbline

\begin{minipage}[t]{0.48\textwidth}
\textbf{\textcolor{blue!75!black}{\small{Gold:}}}
\begin{lstlisting}[
    basicstyle=\ttfamily\footnotesize, 
    backgroundcolor=\color{blue!5!white},
    breaklines=true, 
    columns=fullflexible, 
    frame=single,
    framesep=1mm,
    framerule=0.5pt,
    rulecolor=\color{blue!30!gray!50},
    keepspaces=true,
]
{
  "name": "Giraffe",
  "eatType": "coffee shop",
  "near": "The Rice Boat"
}
\end{lstlisting}
\end{minipage}%
\hfill
\begin{minipage}[t]{0.48\textwidth}
\textbf{\textcolor{red!60!black}{\small{Prediction:}}}
\begin{lstlisting}[
    basicstyle=\ttfamily\footnotesize, 
    backgroundcolor=\color{red!5!white},
    breaklines=true, 
    columns=fullflexible, 
    frame=single,
    framesep=1mm,
    framerule=0.5pt,
    rulecolor=\color{red!30!gray!50},
    keepspaces=true,
]
{
  "name": "The Rice Boat",
  "eatType": "coffee shop",
  "near": "Giraffe"
}
\end{lstlisting}
\end{minipage}
\end{tcolorbox}

\noindent
In such cases, particularly for agent chaining and function calling applications, attributing values to incorrect keys represents a failure to follow user intent and should be penalized.

\section{Dataset Details}
\label{appendix:dataset_details}

\begin{figure}[h!]
    \centering
    \includegraphics[width=\linewidth]{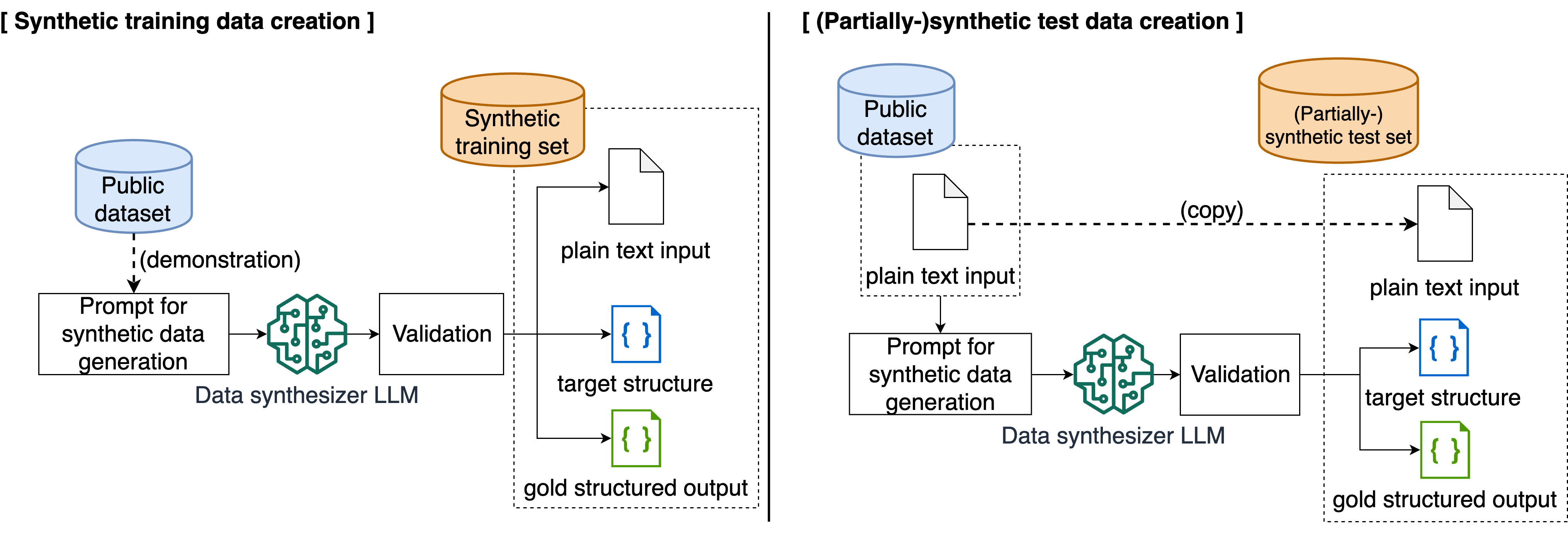}
    \caption{Data curation pipeline for synthetic training data and partially synthetic test data.}
    \label{fig:data_pipeline_appendix}
\end{figure}

We benchmark SPOT on five public datasets, recast into a text-to-JSON formulation.

\subsection{Dataset Curation Details and Examples}
\label{appendix: dataset_curation_details_and_examples}
\paragraph{WebNLG}[License: \href{https://spdx.org/licenses/CC-BY-NC-4.0}{CC BY-NC-SA 4.0}] (DBpedia Triples → Text) In the original dataset, each instance contains data/text pairs where the data is a set of triples extracted from DBpedia, and the text is a verbalisation of these triples. We use the WebNLG v3.0 dataset of ~25K RDF triple sets each paired with 3–7 verbalizations. We converted the triplets into a JSON which described the text / subject.

\begin{lstlisting}[basicstyle=\scriptsize\ttfamily,
  breaklines=true,
  frame=single,
  showstringspaces=false,
  keywordstyle=\color{blue},
  stringstyle=\color{green!60!black}]
{
  "input_text": "The album 1969: The Velvet Underground Live is preceded by the Velvet Underground album Squeeze, which was followed by The Quine Tapes.",
  "json_schema": {
    "type": "object",
    "properties": {
      "category": {
        "type": "string"
      },
      "subject": {
        "type": "string"
      },
      "properties": {
        "type": "object",
        "properties": {
          "precededBy": {
            "type": "string"
          },
          "followedBy": {
            "type": "string"
          }
        },
        "required": ["precededBy", "followedBy"],
        "additionalProperties": false
      }
    },
    "required": ["category", "subject", "properties"],
    "additionalProperties": false
  },
  "gold": {
    "category": "MusicalWork",
    "subject": "Bootleg Series Volume 1: The Quine Tapes",
    "properties": {
      "precededBy": "Squeeze (The Velvet Underground album)",
      "followedBy": "1969: The Velvet Underground Live"
    }
  }
}
\end{lstlisting}

\paragraph{E2E NLG}[License: \href{https://spdx.org/licenses/CC-BY-SA-4.0.html}{CC-BY-4.0}] (Restaurant Attributes → Text) The original dataset is an English benchmark dataset for data-to-text models that verbalize a set of 2-9 key-value attribute pairs in the restaurant domain. Contains restaurant domain descriptions paired with attribute-value structures. This is also one of the dataset that was reversed engineered to text-to-json. 

\begin{lstlisting}[basicstyle=\scriptsize\ttfamily,
  breaklines=true,
  frame=single,
  showstringspaces=false,
  keywordstyle=\color{blue},
  stringstyle=\color{green!60!black}]
{
  "input_text": "The Blue Spice is a pub located by the riverside, near the Rainbow Vegetarian Cafe. It is not child friendly.",
  "json_schema": {
    "type": "object",
    "properties": {
      "name": {
        "type": "string"
      },
      "eatType": {
        "type": "string"
      },
      "area": {
        "type": "string"
      },
      "familyFriendly": {
        "type": "string"
      },
      "near": {
        "type": "string"
      }
    },
    "required": ["name", "eatType", "area", "familyFriendly", "near"],
    "additionalProperties": false
  },
  "gold": {
    "name": "Blue Spice",
    "eatType": "pub",
    "area": "riverside",
    "familyFriendly": "no",
    "near": "Rainbow Vegetarian Cafe"
  }
}
\end{lstlisting}

\paragraph{WikiBio}[License: \href{https://spdx.org/licenses/CC-BY-SA-3.0.html}{CC-BY-SA-3.0}] (Infobox → Intro Paragraph)
The original Dataset contains 728K biographies extracted from Wikipedia containing the first paragraph of the biography and the tabular infobox. This dataset was in json format but not in out desired format to json schema format. We convert the infobox key–value table into a nested JSON schema aligned with first-paragraph content. We randomly sampled 2,500 examples from its test set, providing evaluation on biographical information extraction with diverse schema complexity.
\begin{lstlisting}[basicstyle=\scriptsize\ttfamily,
  breaklines=true,
  frame=single,
  showstringspaces=false,
  keywordstyle=\color{blue},
  stringstyle=\color{green!60!black}]
{
  "input_text": "miroslav popov -lrb- born 14 june 1995 in dvur kralove nad labem -rrb- is a czech grand prix motorcycle racer . he currently races in the fim cev moto2 championship for montaze broz racing team aboard a suter .",
  "json_schema": {
    "type": "object",
    "properties": {
      "birth_date": {
        "type": "string"
      },
      "name": {
        "type": "string"
      }
    },
    "required": ["birth_date", "name"],
    "additionalProperties": false
  },
  "gold": {
    "birth_date": "14 june 1995",
    "name": "miroslav popov"
  }
}
\end{lstlisting}

\paragraph{ToTTo} [License: \href{https://spdx.org/licenses/CC-BY-SA-3.0.html}{CC-BY-SA-3.0}] (Table → Text)  We sampled 500 examples from the challenge split of \href{https://huggingface.co/datasets/GEM/totto}{ToTTo}, a table-to-text dataset where reference texts are annotated based on Wikipedia tables. Since the source tables often contain information beyond the referenced text, we employed LLM annotation to synthesize JSON schema and structured output with further validation and filtering to ensure the structured data strictly adhere to the text content.

\begin{lstlisting}[basicstyle=\scriptsize\ttfamily,
  breaklines=true,
  frame=single,
  showstringspaces=false,
  keywordstyle=\color{blue},
  stringstyle=\color{green!60!black}]
{
  "input_text": "Cypresses are numbered B.152 and date from 1887.",
  "json_schema": {
    "type": "object",
    "properties": {
      "composition": {
        "type": "object",
        "properties": {
          "title": {"type": "string"},
          "catalog_number": {"type": "string"},
          "year": {"type": "integer"}
        },
        "required": ["title", "catalog_number", "year"]
      }
    },
    "required": ["composition"]
  },
  "gold": {
    "composition": {
      "title": "Cypresses",
      "catalog_number": "B.152",
      "year": 1887
    }
  }
}
\end{lstlisting}

\paragraph{HF GitHub Issues} [License: \href{https://www.apache.org/licenses/LICENSE-2.0}{Apache-2.0}] (Issue Text → Structured Issue Report) We sampled 1,000 GitHub issues from \href{https://api.github.com/repos/huggingface/transformers/issues}{Hugging Face transformers repository} up to Feb 2025. For each issue, we combined the title and body text as input, then synthesized the corresponding JSON schema and structured output following the same procedure for ToTTo. The structured outputs capture diverse aspects of software issues including system / error information, reproduction steps, code snippets and expected behaviors, making it an excellent test of model's capability to handle intricate nested structures and technical input contents.

\begin{lstlisting}[basicstyle=\scriptsize\ttfamily,
  breaklines=true,
  frame=single,
  showstringspaces=false,
  keywordstyle=\color{blue},
  stringstyle=\color{green!60!black}]
{
  "input_text": "## qwen2_5_vl processor padding side is wrong.\n### System Info\n\n![Image](https://github.com/user-attachments/assets/6ecbc96d-d34a-4164-903a-0ef65ea65fb0)\n\n![Image](https://github.com/user-attachments/assets/e92f3446-3e81-4887-9f9c-0b5cb3047683)\n\n![Image](https://github.com/user-attachments/assets/8bef88c1-40ba-413b-8444-d018c9691787)\nthe padding side should be left as qwen2 vl do .\n\n### Information\n\n- [ ] The official example scripts\n- [x] My own modified scripts\n\n### Tasks\n\n- [ ] An officially supported task in the `examples` folder (such as GLUE/SQuAD, ...)\n- [x] My own task or dataset (give details below)\n\n### Reproduction\n\nrun conditional generation using qwen2_5_vl  using flash attention 2 .\n\n### Expected behavior\n\n![Image](https://github.com/user-attachments/assets/e92f3446-3e81-4887-9f9c-0b5cb3047683)\n",
  "json_schema": {
    "type": "object",
    "properties": {
      "issue_title": {
        "type": "string"
      },
      "system_info": {
        "type": "array",
        "items": {
          "type": "string"
        }
      },
      "information": {
        "type": "object",
        "properties": {
          "official_example_scripts": {
            "type": "boolean"
          },
          "modified_scripts": {
            "type": "boolean"
          }
        }
      },
      "tasks": {
        "type": "object",
        "properties": {
          "official_task": {
            "type": "boolean"
          },
          "own_task": {
            "type": "boolean"
          }
        }
      },
      "reproduction": {
        "type": "string"
      },
      "expected_behavior": {
        "type": "string"
      }
    }
  },
  "gold": {
    "issue_title": "qwen2_5_vl processor padding side is wrong.",
    "system_info": [
      "https://github.com/user-attachments/assets/6ecbc96d-d34a-4164-903a-0ef65ea65fb0",
      "https://github.com/user-attachments/assets/e92f3446-3e81-4887-9f9c-0b5cb3047683",
      "https://github.com/user-attachments/assets/8bef88c1-40ba-413b-8444-d018c9691787"
    ],
    "information": {
      "official_example_scripts": false,
      "modified_scripts": true
    },
    "tasks": {
      "official_task": false,
      "own_task": true
    },
    "reproduction": "run conditional generation using qwen2_5_vl using flash attention 2.",
    "expected_behavior": "https://github.com/user-attachments/assets/e92f3446-3e81-4887-9f9c-0b5cb3047683"
  }
}
\end{lstlisting}

\subsection{Training Data Diversity Dimensions}
\label{sec:appendix-training-data-details}
\paragraph{Industry Vertical.} Defines the domain context spanning 30 categories such as ``Healthcare'', ``Financial Services'', etc.
\paragraph{JSON Complexity.} Specifies 10 levels of structural complexity from ``Basic'' (3-5 key-value pairs) to ``Comprehensive'' (multiple nested objects)
\paragraph{Text Length Style.} Covering 10 input text length styles from ``Brief Snippets'' (15-30 words) to ``Extended'' (200-300 words)
\paragraph{Genre} Includes 40 text styles (e.g., from news articles, technical reports, etc.)
\paragraph{Text Type.} Specifies 10 types of text structures that are widely present in LLM responses (e.g., ``Bullet points'', ``Code snippets'', ``Dialogue'')

\subsection{Json Complexity Dimensions}
\label{sec:data_complexity_calculation_details}
\subsubsection{Depth ($d$)}
Maximum nesting level of the JSON structure, calculated recursively:
\[ d = \max_{v \in \text{values}} \text{depth}(v) \]
where depth is calculated as:
\[ \text{depth}(v) = \begin{cases}
    0 & \text{if } v \text{ is a primitive value} \\
    1 + \max_{c \in \text{children}(v)} \text{depth}(c) & \text{if } v \text{ is an object or array}
\end{cases} \]

\subsubsection{Number of Keys ($k$)}
Total count of keys at all levels in the JSON structure, calculated as:
\[ k = \sum_{o \in \text{objects}} |\text{keys}(o)| \]
where $|\text{keys}(o)|$ is the number of keys in object $o$, summed across all nested objects.

\subsubsection{Size in Bytes ($s$)}
Size of the JSON string when encoded in UTF-8, calculated as:
\[ s = |\text{UTF-8}(\text{json.dumps}(\text{obj}))| \]
where $|\text{UTF-8}(x)|$ represents the length of string $x$ when encoded in UTF-8.
This has a small weight to avoid dominating the score.

\subsubsection{Number of Elements ($e$)}
Total count of all values in the JSON structure, including primitive values, objects, and arrays. Incremented during traversal:
\[ e = |\{v : v \text{ is any value in the JSON structure}\}| \]

\subsubsection{Cyclomatic Complexity ($c$)}
Measure of structural decision points (branches) in the JSON, calculated as:
\[ c = \sum_{o \in \text{objects}} |\text{keys}(o)| + \sum_{a \in \text{arrays}} [|a| > 0] \]
where:
\begin{itemize}
    \item $|\text{keys}(o)|$ is the number of keys in object $o$
    \item $[|a| > 0]$ is 1 if array $a$ is non-empty, 0 otherwise
\end{itemize}

\subsubsection{Schema Complexity ($sc$)}
Complexity of the JSON structure considering types and nesting. For a value $v$:
\[ sc(v) = \begin{cases}
    1 + \frac{\min(|v|, 100)}{10} & \text{if } v \text{ is a string} \\
    1 + |\text{keys}(v)| + \sum_{k \in \text{keys}(v)} sc(v[k]) & \text{if } v \text{ is an object} \\
    1 + |v| + \begin{cases}
        sc(v[0]) & \text{if homogeneous} \\
        \sum_{i} sc(v[i]) & \text{if heterogeneous}
    \end{cases} & \text{if } v \text{ is an array} \\
    1 & \text{otherwise}
\end{cases} \]
where:
\begin{itemize}
    \item $|v|$ is the length of a string or array, or number of keys in an object
    \item An array is homogeneous if all items have the same type
    \item For homogeneous arrays, complexity is calculated using the first item's schema
\end{itemize}

\subsubsection{Content Complexity ($cc$)}
Complexity of the actual data values. For a value $v$:
\[ cc(v) = \begin{cases}
    \sum_{k \in \text{keys}(v)} cc(v[k]) & \text{if } v \text{ is an object} \\
    \sum_{i} cc(v[i]) & \text{if } v \text{ is an array} \\
    lf + ef + sf + tf & \text{if } v \text{ is a string} \\
    \min(\frac{|\text{digits}(v)|}{5}, 1) & \text{if } v \text{ is a number} \\
    0 & \text{otherwise}
\end{cases} \]
where for strings:
\begin{itemize}
    \item $lf = \min(\frac{|v|}{20}, 5)$ (length factor)
    \item $ef = \frac{|\text{unique}(v)|}{|v|} \cdot 3$ (entropy factor)
    \item $sf = \min(\frac{|\text{special}(v)|}{|v|} \cdot 5, 3)$ (special characters factor)
    \item $tf = 2$ if contains code-like patterns, 0 otherwise (structure factor)
    \item $\text{special}(v)$ counts non-alphanumeric, non-space characters
\end{itemize}

\section{Experiment Details}
\subsection{Training and Inference Hyperparameters}
\label{sec:hyperparameters}

Hyperparameters and configurations used during training and inference are presented in \cref{tab:hyperparams}.
\begin{table*}[h!]
    \centering
    \footnotesize
    \rowcolors{2}{gray!15}{white}
    \resizebox{1\textwidth}{!}{
        \begin{tabular}{p{4cm}|c||p{4cm}|c}
            \hline
            \multicolumn{2}{c||}{\textbf{Training Parameters}} & \multicolumn{2}{c}{\textbf{Inference Parameters}} \\
            \hline
            Learning rate & 1e-5 & Maximum generation length & 2048 \\
            Training epochs & 2 & Temperature & 0 \\
            Per-device batch size & 2 & Top-p & 0.9 \\
            Gradient accumulation & 2 & Top-k & 50 \\
            Context length & 16,384 & Generation timeout & 60 s \\
            Optimizer & paged\_adamw\_32bit & GPU memory utilization & 0.8 \\
            Maximum gradient norm & 0.1 & Tensor parallel size & 1 \\
            \hline
            \multicolumn{4}{l}{\textbf{LoRA Configuration}} \\
            \hline
            Rank (r) & 32 & & \\
            Alpha & 64 & & \\
            Dropout & 0.05 & & \\
            \hline
        \end{tabular}
    }
    \caption{Hyperparameters used in our experiments}
    \label{tab:hyperparams}
\end{table*}

\subsection{Error Analysis}
\label{appendix:error_analysis}

We conducted a detailed analysis of the errors made by different model configurations to better understand their failure modes. Non-fine-tuned models primarily failed by completely ignoring the target schema, either generating free-form text or incorrect JSON structures that did not match the requirements. This matches our expectation that base models lack the specialized knowledge needed to follow complex structural constraints without additional training or guidance.

SFT models showed more sophisticated error patterns, typically involving missing fields, incorrect field types, or hallucinatory content for complex examples. The GitHub Issues dataset accounted for the majority of errors, with failures often occurring in deeply nested structures. We observed that errors frequently appeared at deeper nesting levels (beyond 3-4 levels), suggesting that even fine-tuned models struggle to maintain structural coherence across extended hierarchical dependencies. This points to a fundamental limitation in how standard decoder-only transformer architectures represent and track deeply nested structures during generation.

Constrained decoding approaches (Outlines and XGrammar) guaranteed structural correctness for the examples they could complete but sometimes produced semantically incorrect content, particularly for ambiguous inputs or when multiple valid interpretations were possible. This semantic mismatch manifested in different ways between the two algorithms. XGrammar sometimes over-constrains the model's generative capabilities when the schema is highly structured, particularly affecting the nuanced language expression in fields with longer texts like descriptions. Outlines, with its regex-based validation, occasionally creates token selection pressures that lead to less fluent text in certain fields.

The combined SFT + constrained decoding approach yielded the fewest errors. The few remaining errors were predominantly in the GitHub Issues dataset and typically involved either timeout issues or extreme edge cases where the model struggled to extract the correct content for very complex nested structures. We identified two main error categories in this combined approach: (1) timeout failures when compiling extremely complex schemas with many optional fields and alternatives, and (2) cases where the model must make subjective decisions about how to map ambiguous content to structured fields. 

This error analysis reinforces the complementary nature of SFT and constrained decoding approaches: SFT addresses semantic understanding and content organization, while constrained decoding enforces structural correctness. It also highlights an important direction for future research: developing more efficient algorithms for handling extremely complex schemas and improving models' ability to maintain coherent structural representations across deep hierarchies.

\section{Prompts}

\subsection{Prompts for Training and Evaluation}
\label{appendix:prompt:direct_prompting}
\begin{figure*}[hb!]
\footnotesize
\noindent\fbox{%
    \parbox{\textwidth}{%
        \texttt{Convert the following text into JSON format according to the specified schema. 
Ensure that both keys and values are strings, even for numerical values.}\newline

\texttt{Text: \textcolor{teal}{\{input\_text\}}}\newline

\texttt{Provide your response in the following JSON format: \textcolor{teal}{\{json\_schema\}}}\newline

\texttt{Please output ONLY the JSON structure and extract the attributes only present in the schema.}\newline

\texttt{Output:}

    }%
}%
\caption{Prompt template for direct LLM prompting}
\label{fig:prompt_direct_prompting}
\end{figure*}

\begin{figure*}[hb!]
\footnotesize
\noindent\fbox{%
    \parbox{\textwidth}{%
        \texttt{Be an impartial judge to identify whether the structured data accurately reflect the input text. If the structured data contains anything that unsupported by the `input\_text', return False. If everything can be found or inferred from the input text, return True. Enclose your answer in <validity></validity> xml tags.}\newline

\texttt{<input\_text>\textcolor{teal}{\{input\_text\}}</input\_text>}\newline

\texttt{<structured\_data>\textcolor{teal}{\{gold\}}</structured\_data>}\newline

\texttt{Your answer:}\newline

    }%
}%
\caption{Prompt template for synthetic training data validation}
\label{fig:prompt_data_validation}
\end{figure*}

\label{appendix:prompts}

\subsection{Prompts for Data Generation and Validation}
The prompt template used for synthetic table generation is listed as below. The variables are sampled from different data diversity dimensions as mentioned in \cref{sec:appendix-training-data-details}.
\newline
\newline

\label{appendix:prompt:data_generation_valdiation}
\begin{figure*}[p]
\scriptsize
\noindent\fbox{%
    \parbox{\textwidth}{%
        \texttt{You are an advanced AI assistant specialized in data augmentation for text-to-JSON conversion tasks. Your goal is to generate diverse and high-quality input-output pairs that will be used to train machine learning models for structured information extraction.}\newline
        
\texttt{First, review these examples of input-output pairs:}\newline

\texttt{<examples>}\textcolor{teal}{\{examples\}}\texttt{</examples>}\newline

\texttt{Your task is to generate 3 additional input-output pairs in JSONL format. Each pair should consist of:}\newline
\texttt{1. Input: } \newline
\texttt{   - A text description} \newline
\texttt{   - A desired JSON schema with a brief explanation} \newline
\texttt{2. Output: } \newline
\texttt{   - The text converted into JSON following the given schema} \newline

\texttt{Please adhere to the following specifications:}\newline

\texttt{<industry\_vertical>}
\textcolor{teal}{\{industry\_vertical\}}
\texttt{</industry\_vertical>}

\texttt{<genre>}
\textcolor{teal}{\{genre\}}
\texttt{</genre>}

\texttt{<json\_complexity\_description>}
\textcolor{teal}{\{json\_complexity\_description\}}
\texttt{</json\_complexity\_description>}

\texttt{<text\_length\_style>}
\textcolor{teal}{\{text\_length\_style\}}
\texttt{</text\_length\_style>}

\texttt{<text\_type>}
\textcolor{teal}{\{text\_type\}}
\texttt{</text\_type>}\newline

\texttt{Before generating each pair, wrap your thought process in <planning> tags. Consider the following:}\newline
    
\texttt{1. Industry relevance: How can you make the text diverse and representative of the specified industry - *\textcolor{teal}{\{industry\_vertical\}}*?} \newline
\texttt{   - List 1-5 key topics or scenarios relevant to the industry.} \newline
\texttt{   - Consider how these topics can be incorporated into the text descriptions.} \newline
\texttt{2. JSON schema complexity: How can you adhere to the complexity level of the JSON schema - *\textcolor{teal}{\{json\_complexity\_description\}}*, while staying within the other given constraints?} \newline
\texttt{   - Brainstorm 1-5 different schema structures of such complexity.} \newline
\texttt{   - Ensure each schema adheres to the complexity description and other constraints provided.} \newline
\texttt{3. Text length adherence: How can you ensure the text length adheres to the specified style - *\textcolor{teal}{\{text\_length\_style\}}*?} \newline
\texttt{   - Outline a strategy for maintaining consistent text length across all pairs.} \newline
\texttt{   - Consider using a word count check for each generated text.} \newline
\texttt{4. Unique aspects and edge cases: What unique aspects or edge cases can you incorporate to make the training data more robust?} \newline
\texttt{   - List 2-3 potential edge cases or unusual scenarios relevant to the industry.} \newline
\texttt{   - Plan how to integrate these into some of the pairs.
} \newline
\texttt{5. Diversity tracking: How will you ensure diversity across all 5 pairs?} \newline
\texttt{   - Create a simple tracking system to ensure you're varying topics, schema complexity, and text length across the pairs.} \newline
\texttt{   - Number each pair as you plan it (1/5, 2/5, etc.) to keep track of your progress.} \newline
\texttt{6. JSON content alignment: How will you ensure the gold JSON only contains information present in the input text?} \newline
\texttt{   - Implement a strict check to verify that every piece of information in the gold JSON can be directly traced back to the input text.} \newline
\texttt{   - Plan a review process to eliminate any potential extra content in the gold JSON that is not explicitly stated in the input text.} \newline
\texttt{7. Genre: Ensure that the generated text adheres to the specified genre *\textcolor{teal}{\{genre\}}*. This will influence the tone, style, and content of the `input\_text`, but does not affect the JSON structure.} \newline
\texttt{8. Text Type: Generate text that aligns with the specified text type *\textcolor{teal}{\{text\_type\}}*. This will determine the format, structure, or purpose of the `input\_text` you create, separate from the JSON output.} \newline
\texttt{9. Follow the format in the example below for your `input\_text`, 'json\_schema` and `gold`:} \newline

\texttt{\{\{
    "input\_text": "Acme Motors, a major automaker has announced plans to build a new electric vehicle manufacturing plant in Greenville, South Carolina. The state-of-the-art facility will produce the company's latest line of battery-powered cars and SUVs, with an initial annual capacity of 200,000 units.",
    "json\_schema": \{\{
        "type": "object",
        "properties": \{\{
            "company": \{\{
                "type": "string"
            \}\},
            "location": \{\{
                "type": "string"
            \}\},
            "production\_capacity": \{\{
                "type": "number"
            \}\}
        \}\}
    \}\},
    "gold": \{\{
        "company": "Acme Motors",
        "location": "Greenville, South Carolina",
        "production\_capacity": 200000
    \}\}
\}\}}\newline

\texttt{Notes:} \newline
        \texttt{- "input\_text": Contains a string value, represents the raw text input that needs to be processed}\newline
        \texttt{In this case, it's a news-like paragraph about a company announcement.}\newline
        \texttt{- "json\_schema": Defines the structure of the expected output. must include "type" and "properties", defined as below:}\newline
        \texttt{"type": "object" - specifies that the output should be a JSON object}\newline
        \texttt{"properties" - defines the expected fields:}\newline
        \texttt{   - "company": expects a string value}\newline
        \texttt{   - "location": expects a string value}\newline
        \texttt{   - "production\_capacity": expects a number value}\newline
        \texttt{Each property has its own type definition}\newline
        \texttt{- "gold": Contains the correct/expected output that matches the schema. Has the exact same structure as defined in json\_schema. Contains the actual values extracted from the input\_text:}\newline
        \texttt{   - "company": contains the company name}\newline
        \texttt{   - "location": contains the city and state}\newline
        \texttt{   - "production\_capacity": contains the numerical value}\newline
        \texttt{Values must match the types specified in the schema (strings for company and location, number for production\_capacity)}\newline
        \texttt{- *IMPORTANT* Do not replicate any of the content in the given example, it's just used as a reference for a specified answer structure.}\newline
        
        \texttt{After your planning process, generate the input-output pair and format it in JSONL. Here's an example of the expected format:}\newline
        
        \texttt{\{\{"input\_text": "Your generated text here", "json\_schema": \{\{"Your": "JSON", "schema": "here"\}\}, "gold": \{\{"Your": "gold", "JSON": "here"\}\}\}}\newline
        
        \texttt{Remember to generate 5 unique and diverse pairs, each following this format. Ensure that each pair adheres to the industry vertical, complexity description, and text length style specified above. EACH response should be enclosed in <JSONL> </JSONL> XML tags.}

    }%
}%

\caption{Prompt template synthetic training data generation}
\label{fig:prompt_synthetic_training_data}
\end{figure*}

\end{document}